\documentclass[10pt,twocolumn,letterpaper]{article}

\usepackage[pagenumbers]{cvpr} 
\usepackage{multirow}

\usepackage{tikz}
\usetikzlibrary{arrows.meta}  
\usepackage{graphicx}
\usepackage{pifont}

\usepackage{algorithm}
\usepackage{algorithmic}
\usepackage{float}

\definecolor{cvprblue}{rgb}{0.21,0.49,0.74}
\usepackage[pagebackref,breaklinks,colorlinks,allcolors=cvprblue]{hyperref}
\usepackage[table]{xcolor}  
\usepackage{colortbl}
\usepackage{pgfplots}
\usepackage{tikz}
\usepackage[table]{xcolor}
\pgfplotsset{compat=1.17}
\def\paperID{9131} 
\def\confName{CVPR}
\def\confYear{2026}

\title{Progressive Supernet Training for Efficient Visual Autoregressive Modeling}

\author{%
    Xiaoyue Chen\textsuperscript{\rm 1*},\;
    Yuling Shi\textsuperscript{\rm 2*},\;
    Kaiyuan Li\textsuperscript{\rm 1*}, \;
    Huandong Wang\textsuperscript{\rm 1}
    \\
    Yong Li\textsuperscript{\rm 1},\;
    Xiaodong Gu\textsuperscript{\rm 2},\;
    Xinlei Chen\textsuperscript{\rm 1\dag},\;
    Mingbao Lin\textsuperscript{\rm 3\dag}
    \thanks{*Equal contribution. \dag Co-corresponding authors.}
    \\
    \textsuperscript{\rm 1}Tsinghua University, China \,
    \textsuperscript{\rm 2}Shanghai Jiao Tong University, China \,
    \textsuperscript{\rm 3}Rakuten, Singapore \\
    {\tt\small chenxiao24@mails.tsinghua.edu.cn\, yuling.shi@sjtu.edu.cn \, likaiyua23@mails.tsinghua.edu.cn}\\ 
    {\tt\small wanghuandong@tsinghua.edu.cn, liyong@tsinghua.edu.cn\, xiaodong.gu@sjtu.edu.cn@sjtu.edu.cn} \\
    {\tt\small chen.xinlei@sz.tsinghua.edu.cn\, linmb001@outlook.com}
}

\begin{document}

\maketitle

\begin{abstract}
Visual Auto-Regressive (VAR) models significantly reduce inference steps through the “next-scale” prediction paradigm. However, progressive multi-scale generation incurs substantial memory overhead due to cumulative KV caching, limiting practical deployment.
We observe a \textit{scale-depth asymmetric dependency} in VAR: early scales exhibit extreme sensitivity to network depth, while later scales remain robust to depth reduction. Inspired by this, we propose \textbf{VARiant}: by equidistant sampling, we select multiple subnets ranging from 16 to 2 layers from the original 30-layer VAR-d30 network. Early scales are processed by the full network, while later scales utilize subnet. Subnet and the full network share weights, enabling flexible depth adjustment within a single model. 
However, weight sharing between subnet and the entire network can lead to optimization conflicts. To address this, we propose a progressive training strategy that breaks through the Pareto frontier of generation quality for both subnets and the full network under fixed-ratio training, achieving joint optimality.
Experiments on ImageNet demonstrate  that, compared to the pretrained VAR-d30 (FID 1.95), VARiant-d16 and VARiant-d8 achieve nearly equivalent quality (FID 2.05/2.12) while reducing memory consumption by 40-65\%. VARiant-d2 achieves 3.5$\times$ speedup and 80\% memory reduction at moderate quality cost (FID 2.97). In terms of deployment, VARiant's single-model architecture supports zero-cost runtime depth switching and provides flexible deployment options from high quality to extreme efficiency, catering to diverse application scenarios.
Our project is available at \url{https://github.com/Nola-chen/VARiant}
\end{abstract}

\section{Introduction}

Autoregressive (AR) architectures have demonstrated outstanding performance in natural language processing~\cite{qwen2025qwen25technicalreport,deepseekai2025deepseekr1incentivizingreasoningcapability,grattafiori2024llama3herdmodels} and image understanding~\cite{bai2025qwen25vltechnicalreport,guo2025seed15vltechnicalreport}, while also driving the expansion of research in image synthesis~\cite{liu2024lumina,wu2025janus,wu2024vila}. However, traditional next-token prediction methods~\cite{lee2022autoregressive,razavi2019generating,yu2021vector} suffer from suboptimal visual quality and slow generation due to discrete tokenization and sequential sampling. To address this, Visual Autoregressive (VAR)~\cite{tian2024visual} introduces a next-scale prediction paradigm that generates images from coarse to fine, improving both quality and speed through parallel generation across spatial scales.

Despite its promising performance, VAR introduces a notable challenge in memory efficiency during inference. Generating finer-scale representations requires retaining all previously generated tokens across scales, leading to significantly higher memory consumption than standard autoregressive models. Recent works have explored strategies to mitigate this bottleneck, including step-level distillation~\cite{liu2024distilled}, token-level compression~\cite{guo2025fastvar}, KV-cache optimization~\cite{qin2025head}, and multi-model layer-wise scheduling~\cite{chen2025collaborative}. However, they either compromise generation fidelity or introduce system overhead and deployment complexity.

Through an in-depth analysis of VAR's generation mechanism in Sec.\,\ref{subsub:observation}, we observe a \textit{scale-depth asymmetric dependency}: early scales exhibit highly sensitive to model depth (50\% depth subnet lead to FID degradation exceeding 20), while later scales exhibit robustness to depth (FID differences less than 4). While existing multi-model collaboration methods (\emph{e.g.}, CoDe~\cite{chen2025collaborative}) can leverage this property for acceleration, they require deploying multiple independent models. Our goal is to achieve scale-wise flexible depth adjustment within a single model, thereby avoiding the system complexity introduced by multi-model deployment.

We propose \textbf{VARiant}, a unified supernet framework that supports multiple depth configurations within a single model. Subnets are selected from the full network via equidistant sampling, with early scales processed by the full network and later scales by shallow subnets. This design provides two advantages: \textbf{(1) Implicit knowledge transfer}---subnet layers share weights with the full network and undergo collaborative training; \textbf{(2) Cross-scale gradient propagation}---skipped layers still receive gradient updates through early scales.

However, weight sharing introduces \textbf{optimization conflicts}: training only subnets degrades full-network performance, while training only the full network hinders effective subnetwork learning. To address this, we propose a \textbf{dynamic-ratio progressive training} strategy: the initial stage samples subnets at low probability (20\%, empirically optimal) to establish a parameter foundation; the intermediate stage gradually increases the sampling ratio for smooth transition, and the final stage focuses on subnet optimization. This progressive design successfully breaks through the Pareto frontier limitations of fixed-ratio training, ensuring both the full network and subnets achieve optimal performance simultaneously.

Experiments on ImageNet 256$\times$256~\cite{deng2009imagenet} demonstrate that VARiant achieves flexible quality-efficiency trade-offs by adjusting subnet depth within a single model. The recommended configuration (16-layer subnet) achieves 1.7$\times$ inference acceleration and 44\% memory savings, with FID increasing only from 1.96 to 2.05; shallower 8-layer and 2-layer subnets achieve 2.6$\times$ and 3.5$\times$ speedups respectively, with 65\% and 80\% memory savings, maintaining FID of 2.15 and 2.67, preserving usable generation quality even in extreme efficiency scenarios. In terms of deployment, VARiant's single-model design supports zero-cost runtime depth switching, flexibly adapting to diverse deployment scenarios ranging from high quality to extreme efficiency.

\begin{figure*}[t]
\centering
\includegraphics[width=1.9 \columnwidth]{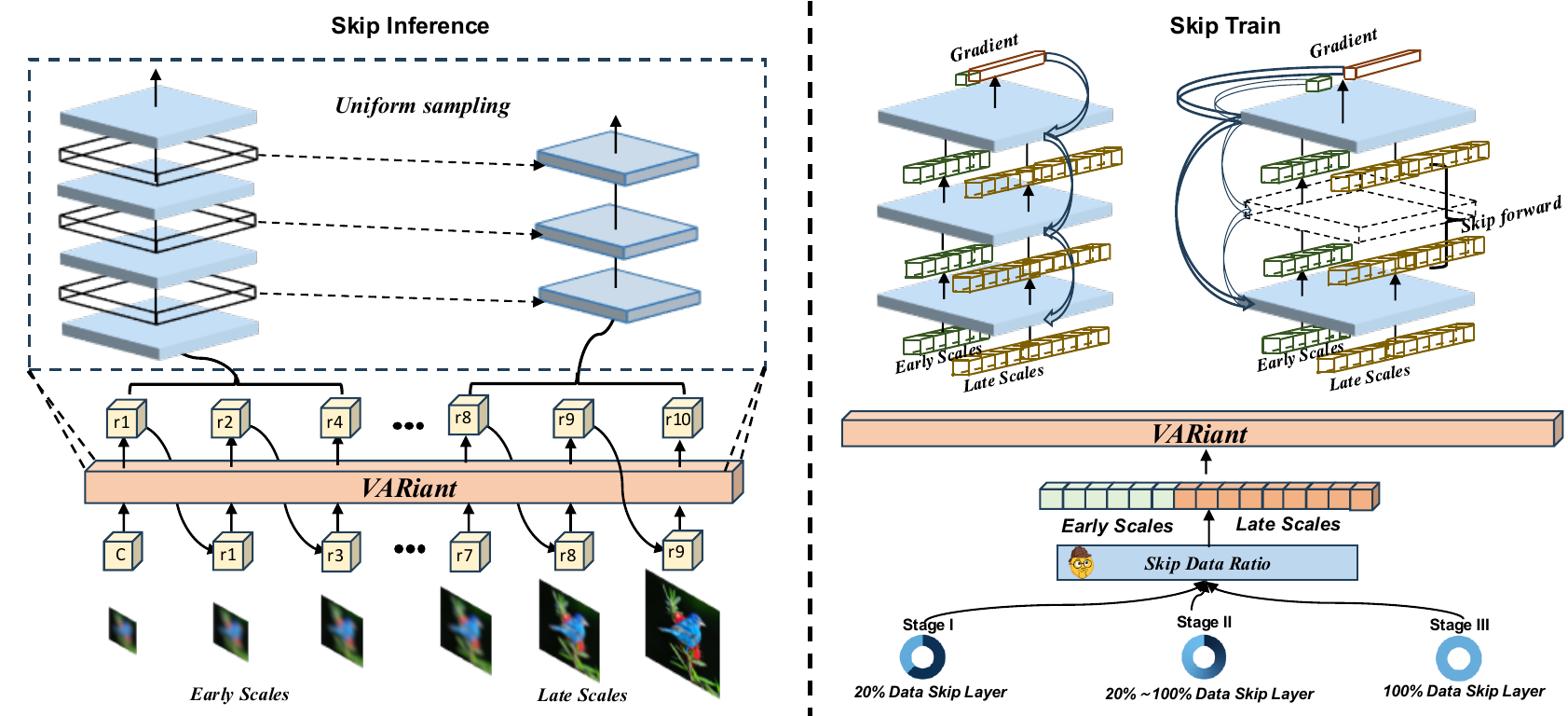}
\caption{VARiant inference and training framework.}
\label{fig:method_main}
\end{figure*}

\section{Related Work}

\subsection{Autoregressive Visual Generation}

The successful application of autoregressive (AR) architectures in large-scale language modeling~\cite{achiam2023gpt, devlin2019bert,chang2023muse} has spurred research into their use for visual synthesis. Traditional visual autoregressive models~\cite{lee2022autoregressive,razavi2019generating,yu2021vector} quantize images into discrete token sequences~\cite{van2017neural,esser2021taming} and synthesize them token-by-token. However, this discretization and serial decoding limit both generative fidelity and sampling efficiency. VAR~\cite{tian2024visual} reformulates autoregressive decoding as a next-scale paradigm, enabling coarse-to-fine generation with hierarchical parallelism, significantly improving image quality and inference latency.

Multiple extensions have emerged from VAR:  ControlVAR~\cite{li2024controlvar, li2024controlar} introduces pixel-level controllability, SAR3D~\cite{chen2024sar3d} and VAT~\cite{zhang20243drepresentation512bytevariationaltokenizer} extend to 3D object synthesis, while VARSR~\cite{qu2025visual} and Infinity~\cite{han2025infinity} apply to high-resolution super-resolution. However, VAR still faces a critical challenge: KV cache accumulation grows quadratically with resolution during inference, creating a memory bottleneck that limits practical deployment.

\subsection{Efficient Autoregressive Generation}

To address the computational challenges of VAR models, existing acceleration strategies fall into three categories: step-level distillation, token-level compression, and multi-model hybrid scheduling.
Step-level distillation aims to reduce the number of generation steps. Distilled Decoding~\cite{liu2024distilled} compresses multi-step VAR generation into one or two steps, achieving approximately 3$\times$ speedup, but this aggressive compression results in substantial quality degradation.
Token-level or cache-level compression methods reduce memory overhead by selectively retaining important tokens or KV cache entries. FastVAR~\cite{guo2025fastvar} and HACK~\cite{qin2025head} achieve 50\%-70\% KV cache compression. While these methods preserve generation quality better, they require fine-grained token-level operations that complicate implementation and reduce deployment flexibility.
Multi-model hybrid scheduling provides a complementary strategy by adjusting computational depth. CoDe~\cite{chen2025collaborative} employs a collaborative framework where a small auxiliary model and a large model process different scales, but requires deploying two independent models, increasing system complexity and memory consumption.

In contrast, our supernet-based approach supports dynamic depth adjustment within a single model, eliminating the need for multiple model instances or complex token operations. This provides a more elegant, flexible, and easily deployable VAR acceleration solution while maintaining competitive generation quality.

\section{Methodology}

\subsection{Preliminary: Visual Autoregressive Modeling}

Visual autoregressive (VAR) modeling~\cite{tian2024visual} reformulates traditional autoregressive generation by shifting from ``next-token'' prediction to ``next-scale'' prediction. Given an image feature map $\mathbf{f} \in \mathbb{R}^{h \times w \times C}$, VAR quantizes it into $K$ multi-scale token maps $\mathcal{R} = (r_1, r_2, \ldots, r_K)$ at progressively finer resolutions. The joint probability distribution is factorized as:
\begin{equation}
p(r_1, r_2, \ldots, r_K) = \prod_{k=1}^{K} p(r_k \mid r_1, r_2, \ldots, r_{k-1}),
\end{equation}
where each token map $r_k \in [V]^{h_k \times w_k}$ consists of $h_k \times w_k$ discrete tokens from a vocabulary of size $V$ at scale $k$. 
At each autoregressive step $k$, the model concurrently predicts all $h_k \times w_k$ tokens in $r_k$ based on prior scale conditions.
VAR employs a unified transformer with $D$ layers to process all scales, where all tokens from different scales share the same network parameters.

\subsection{VARiant Supernet}

In this section, we conduct exploratory experiments on ImageNet–256$\times$256, using VAR-d30~\cite{tian2024visual} as the full-depth baseline, results of which directly inform the design of our VARiant method. Then, we propose our VARiant with its inference and training framework provided in Figure\,\ref{fig:method_main}.

\subsubsection{Observation: Scale-Depth Asymmetric Dependency}\label{subsub:observation}

To explore efficient deployment strategies for VAR, we systematically investigate how network depth affects generation quality across different scales. 

Table\,\ref{tab:scale_depth_sensitivity} summarizes the findings.
When the shallow subnet (50\% layers) is deployed on the low-resolution scales ($r_1$-$r_3$), FID jumps from the full-model 1.95 to 12.91 (+10.95), indicating a near-total loss of global semantics. 
Applying the same shallow subnet to the mid-resolution scales ($r_4$-$r_6$) yields FID = 8.5, while restricting it to the high-resolution scales ($r_7$-$r_{10}$) gives FID = 5.42---only +3.47---while reducing layer-wise FLOPs by 46.7\% for 87\% of the overall inference latency.
This striking \textbf{scale-depth asymmetric dependency} shows that the low-resolution stages, which establish global layout and semantic structure, critically require the representational capacity of deep networks, whereas the high-resolution stages, which refine local textures, are robust to depth reduction. 



\begin{table}[h]
\centering
\caption{Impact of subnet application on different scales. Applying subnets to early scales causes severe quality degradation, while applying to later scales preserves quality.}
\label{tab:scale_depth_sensitivity}
\resizebox{\columnwidth}{!}{
\begin{tabular}{lccc}
\toprule
\textbf{Strategy} & \textbf{$d$=30 scales} & \textbf{$d$=16 scales} & \textbf{Final FID} \\
\midrule
Full depth & $r_1$-$r_{10}$ & None & 1.95 \\
\midrule
Early subnet & $r_4$-$r_{10}$ & $r_1$-$r_3$ & \textcolor{red}{12.91} \\
Mid subnet & $r_1$-$r_3$, $r_7$-$r_{10}$ & $r_4$-$r_6$ & 8.5 \\
\rowcolor{green!20}
\textbf{Late subnet (Ours)} & \textbf{$r_1$-$r_6$} & \textbf{$r_7$-$r_{10}$} & \textcolor{green}{\textbf{5.42}} \\
\bottomrule
\end{tabular}}
\end{table}

\subsubsection{Architecture Design: Supernet with Shared Weights}

Motivated by the scale-depth asymmetric dependency, we construct a unified supernet that supports multiple depth configurations within one set of parameters. Depth becomes a real-time adjustable hyper-parameter, eliminating the need to store or load multiple models.

\textbf{Equidistant Layer Selection.} 
Given full depth \(D\) and a target subnet depth \(d\) (\emph{e.g.}, \(d = \frac{1}{2}D\) or \(d = \frac{1}{4}D\)), we obtain the active-layer index set by equidistant sampling while always retaining the first and last layers:
\begin{equation}\label{eq:selection}
\mathcal{I}_d = \left\{ \left\lfloor \frac{i \cdot (D-1)}{d-1} \right\rfloor \;\bigg|\; i = 0, 1, \ldots, d-1 \right\}.
\end{equation}

Consequently $\mathcal{I}_d$ is nested: $\mathcal{I}_{0.25D} \subset \mathcal{I}_{0.5D} \subset \{0, ..., D-1\}$, which maximizes parameter sharing and knowledge transfer across depths.

\begin{figure*}[!t] 
\centering 
\includegraphics[width=1.0 \linewidth]{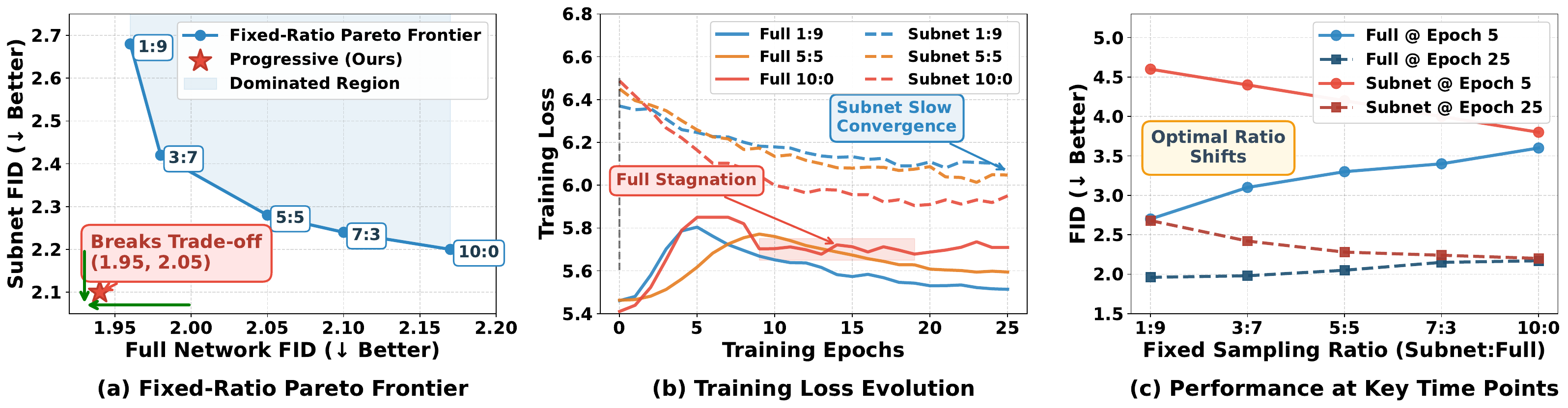}  %
\caption{Fixed-ratio training exhibits (a) Pareto trade-offs, (b) optimization conflicts at extreme ratios, and (c) time-varying optimal ratios, motivating our progressive training strategy.}
\label{fig:optimization_analysis} 
\end{figure*}

\textbf{Cross-Scale Depth Allocation.} 
Inspired by the asymmetric dependency phenomenon, we split the $K$-scale generation pipeline into two functional zones.
\begin{itemize}[leftmargin=*,nosep]
\item \textbf{Bridge Zone} ($r_1$--$r_N$): always executed with the full $D$ layers to protect global semantics.
\item \textbf{Flexible Zone} ($r_{N+1}$--$r_K$): runtime choice among a discrete depth set $\mathcal{I}_d$.
\end{itemize}

Formally, the active layer set at step $k$ is
\begin{equation}\label{eq:active}
\mathcal{I}_k = 
\begin{cases}
\{0, 1, \ldots, D-1\}, & \text{if } k \leq N \text{ (Bridge Zone)}, \\
\mathcal{I}_d, & \text{if } k > N \text{ (Flexible Zone)},
\end{cases}
\end{equation}
where $d$ can be switched on-the-fly to meet latency, memory or quality budgets.

By this way, we provide advantages including
1. \emph{Single-model store}: A single model file eliminates version conflicts and storage overhead.
2. \emph{Zero loading latency}: Dynamic depth switching via layer indexing without reloading.
3. \emph{Excellent compatibility}: The standard Transformer architecture ensures cross-platform compatibility.

\subsection{Progressive Training Strategy}

\subsubsection{Observation: Training Conflicts and Fixed-Ratio Limitations}

Although the unified supernet architecture supports depth configuration during inference, a key question remains: how to train so that \emph{both} the full-depth network and any shallow subnet reach their respective optima?
We first examine the simplest strategy---fixed-ratio joint training---where the shallow configuration $\mathcal{I}_d$ is sampled with constant probability $p$ and the full depth with $1-p$.

\textbf{Pareto Frontier of Fixed Ratios.} Figure\,\ref{fig:optimization_analysis}(a) shows the bi-objective space (full-network FID \emph{vs.} subnet FID) obtained by varying $p$ from 0.1 to 1.0.
No single $p$ simultaneously optimizes both:
$p=0.1$ achieves the best full-network FID (1.96) but degrades the subnet to $2.68$, while $p=1.0$ improves the subnet to $2.15$ but raises the full-network FID to $2.32$.
The smooth Pareto front confirms that any constant ratio is a compromise.

\textbf{Gradient-Starvation Dynamics.}
Figure\,\ref{fig:optimization_analysis}(b) shows the training dynamics under extreme ratios.
With $p=1.0$, the full network stagnates after epoch 8 (loss drops only from 5.75 to 5.72), because layers unused by the subnet receive gradients solely from the Bridge Zone ($\approx$30\% tokens), insufficient for effective updates.
Conversely, $p=0.1$ slows subnet convergence: the shallow path is activated too rarely to specialize its weights, leaving the subnet loss at a higher plateau (6.08).

\textbf{Stage-Dependent Requirements.}
Figure\,\ref{fig:optimization_analysis}(c) compares the same ratios at epoch 5 and epoch 25.
Early in training, the full network is still inaccurate (FID 3.6-4.6); high $1-p$ is necessary to build a parameter foundation for \emph{all} layers.
Late in training, the full network has converged (FID $\le$2.2); high $p$ allows the layers retained by $\mathcal{I}_d$ to specialize for shallow topology.
Fixed ratios unsatisfies these demands, motivating a \textbf{dynamic sampling schedule} that emphasizes full-depth updates early and subnet updates late.

\subsubsection{Dynamic Ratio Training}
\begin{figure}[!t] 
\centering 
\includegraphics[width=\columnwidth]{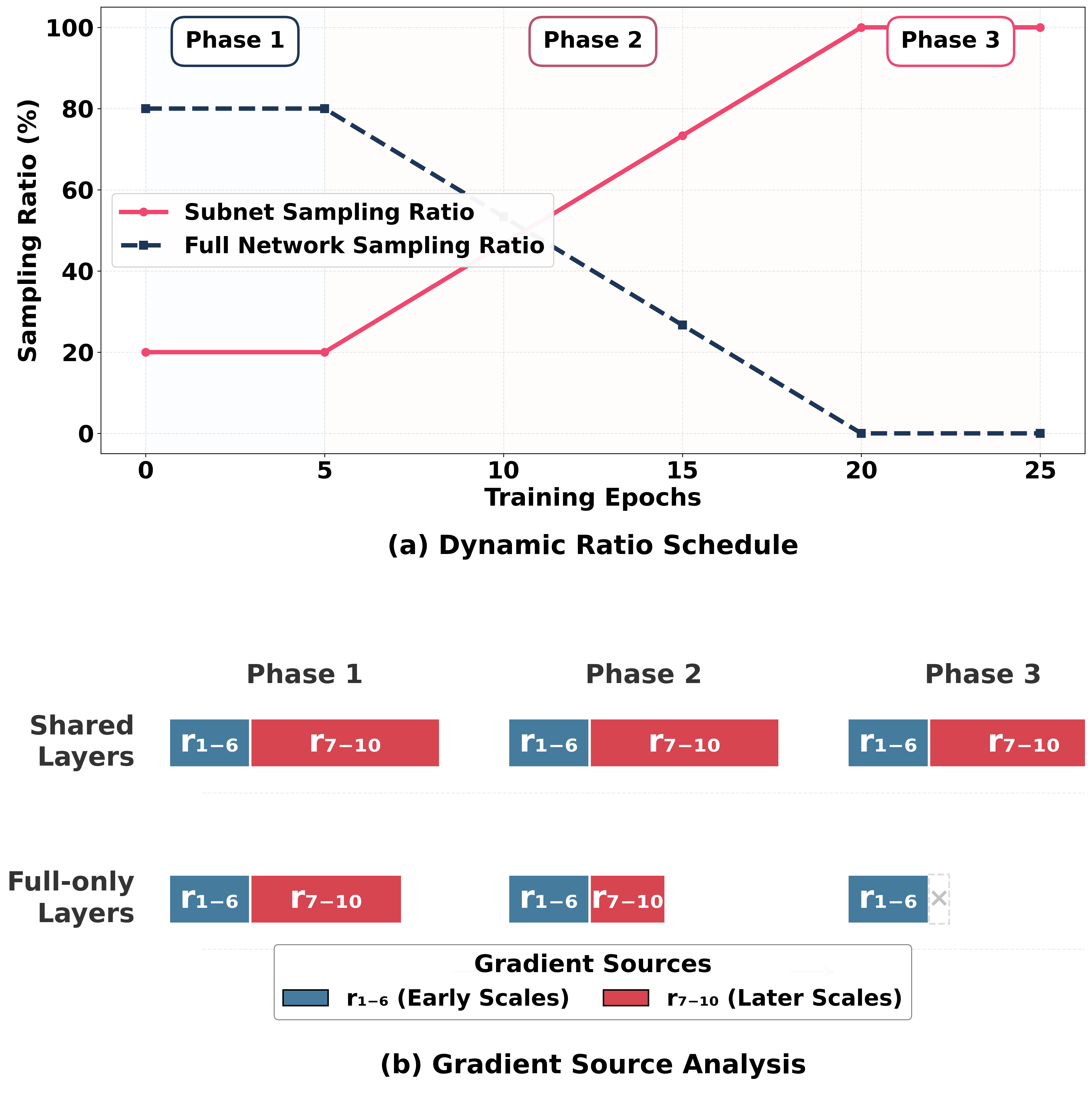}  %
\caption{Progressive training strategy. (a) Dynamic sampling ratio schedule across three training phases. (b) Gradient source analysis showing the transition from joint optimization to subnet-focused refinement through a stable gradient bridge.}
\label{fig:training_strategy} 
\end{figure}

We translate stage-specific insights into a three-phase training plan that redistributes gradient flow along the timeline by continuously adjusting the sampling ratio $\rho = \text{subnetwork}{:}\text{full network}$; see Figure\,\ref{fig:training_strategy}(a) for details.

During training, model layers are categorized into two types: \textbf{Shared Layers} are used by both subnet and full network (\emph{e.g.}, the 16 layers selected by subnet), while \textbf{Full-only Layers} are used exclusively by the full network and skipped by subnet (\emph{e.g.}, the remaining 14 layers).

\textbf{Phase 1: Joint Training ($[0, E_1]$, $\rho_1=2:8$).} Early training employs small-ratio joint training, with the Flexible Zone using subnet configuration $\mathcal{I}_d$ with 20\% probability. The training objective is:
\begin{equation}\label{eq:phase1_loss}
\mathcal{L} = \sum_{k=1}^{K} \text{CE}\left(p_\theta(r_k \mid r_{<k}, \mathcal{I}_k), r_k^*\right).
\end{equation}

As shown in Figure\,\ref{fig:training_strategy}(b), during this phase, \textbf{Shared Layers} receive gradients from both the Bridge Zone ($r_1$-$r_6$, blue) and the Flexible Zone ($r_7$-$r_{10}$, red); More importantly, \textbf{Full-only Layers}, due to the high probability of full-network sampling in the Flexible Zone, also receive sufficient gradient signals from both zones. This ensures all layers (including those not selected by the subnet) establish a strong parameter foundation.

\textbf{Phase 2: Progressive Transition ($[E_1+1, E_2]$, $\rho_1 \to \rho_2$).} To avoid optimization stagnation from abruptly switching to large ratios, we use a progressive transition where the subnet sampling probability increases linearly:
\begin{equation}
p(\text{ep}) = 0.2 + 0.8 \cdot \frac{\text{ep} - E_1}{E_2 - E_1}.
\end{equation}

Figure\,\ref{fig:training_strategy}(b) shows that as the subnet sampling ratio increases, \textbf{the gradient contribution from the Flexible Zone(red portion) gradually diminishes}, while the gradient from the Bridge Zone (blue portion) remains stable. This continuous adjustment enables model parameters to gradually adapt to the new gradient distribution, avoiding instability caused by abrupt changes.

\textbf{Phase 3: Subnet Refinement ($[E_2+1, E]$, $\rho_2=10:0$).} Later training focuses on refining the subnet configuration, with the Flexible Zone consistently using $\mathcal{I}_d$. As shown in Figure\,\ref{fig:training_strategy}(b), \textbf{Full-only Layers completely lose gradient support from the Flexible Zone} (red portion disappears), relying solely on gradients from the Bridge Zone (blue portion) for optimization. However, the strong parameter foundation established in Phases 1-2 enables this partial gradient to maintain full-network performance while allocating primary computational resources to subnet specialization.

This strategy successfully resolves the limitations of fixed ratios. By dynamically adjusting gradient allocation over time---providing sufficient gradients to all layers during early training to avoid optimization stagnation, and gradually shifting training resources to subnet specialization in later training---we break through the Pareto frontier of fixed ratios. Figure\,\ref{fig:optimization_analysis}(a) shows the Progressive method (red star) achieves optimality for both configurations.

\begin{table*}[t]
\centering
\caption{Quantitative assessment of the efficiency-quality trade-off across various methods. Inference efficiency evaluated with batch size 64 on NVIDIA L20 GPU, latency excluding VQVAE's shared cost.}
\label{tab:main_results}
\small
\setlength{\tabcolsep}{3.5pt}
\begin{tabular}{l|cccccc|cccc}
\toprule
\multirow{2}{*}{\textbf{Method}} & \multicolumn{6}{c|}{\textbf{Inference Efficiency}} & \multicolumn{4}{c}{\textbf{Generation Quality}} \\
\cmidrule(lr){2-7} \cmidrule(lr){8-11}
& \textbf{\#Steps} & \textbf{Speedup↑} & \textbf{Latency↓} & \textbf{Mem↓} & \textbf{KV Cache↓} & \textbf{\#Param} & \textbf{FID↓} & \textbf{IS↑} & \textbf{Prec↑} & \textbf{Rec↑} \\
\midrule
DiT-XL/2 & 50 & -- & 19.20s & -- & -- & 675M & 2.26 & 239 & 0.80 & 0.60 \\
LlamaGen-XXL & 384 & -- & 74.27s & -- & -- & 1.4B & 2.34 & 254 & 0.80 & 0.59 \\
\midrule
VAR-d30 & 10 & 1.0× & 3.62s & 39265MB & 28677MB & 2.0B & 1.95 & 301 & 0.81 & 0.59 \\
VAR-CoDe & 6+4 & 2.9× & 1.27s & 19943MB & 8156MB & 2.0+0.3B & 2.27 & 297 & 0.82 & 0.58 \\
\midrule
VARiant-d16 & 6+4 & 1.7× & 2.12s & 28644MB & 16092MB & 2.0B & 2.05 & 314 & 0.81 & 0.59 \\
VARiant-d8 & 6+4 & 2.6× & 1.40s & 20759MB & 9465MB & 2.0B & 2.12 & 306 & 0.80 & 0.58 \\
VARiant-d4 & 6+4 & 3.0× & 1.21s & 19582MB & 7372MB & 2.0B & 2.28 & 296 & 0.78 & 0.56 \\
\rowcolor{blue!5} \textbf{VARiant-d2} & \textbf{6+4} & \textbf{3.5×} & \textbf{1.03s} & \textbf{18869MB} & \textbf{5495MB} & \textbf{2.0B} & \textbf{2.97} & \textbf{276} & \textbf{0.75} & \textbf{0.53} \\
\bottomrule
\end{tabular}
\end{table*}

\section{Experiments}

\subsection{Experimental Setup}

\textbf{Dataset and Task.} 
We evaluate our method on ImageNet-1K~\cite{deng2009imagenet}, class-conditional generation at 256×256 resolution.

\textbf{Model Configuration.} 
We employ the pre-trained VAR-d30~\cite{tian2024visual} (30 transformer layers) as our base model. Through supernet training, we obtain five depth configurations: 2, 4, 8, 16, and 30 layers (full network).

\textbf{Training Strategy.} We adopt a three-stage dynamic-ratio progressive training approach over 25--35 epochs:
\begin{itemize}
    \item \textit{Stage 1 (Joint Training, 5 epochs)}: Subnet and full network sampling ratio of 2:8.
    \item \textit{Stage 2 (Progressive Transition, 15 epochs)}: Sampling ratio linearly transitions from 2:8 to 10:0.
    \item \textit{Stage 3 (Subnet Refinement, 5--15 epochs)}: Only subnet training (ratio 10:0). The duration is adaptive based on subnet convergence—shallower subnets typically require longer refinement to achieve optimal performance.
\end{itemize}

We use the AdamW optimizer~\cite{loshchilov2017decoupled} with a learning rate of $1 \times 10^{-6}$ and a batch size of 1024. We train our VARiant supernetwork on 8 NVIDIA H100 GPUs.

\textbf{Evaluation Setup.} 
Quality metrics include FID, Inception Score (IS), Precision, and Recall. Efficiency metrics include inference latency, memory consumption, and parameter count. Sampling configuration: top-k=900, top-p=0.96. All efficiency tests are conducted on a single NVIDIA L20 GPU, with timing excluding the VQVAE decoder.

\noindent\textbf{Comparison Methods.} 
We compare against: (1) diffusion models (DiT); (2) traditional autoregressive models (LlamaGen); (3) original VAR\_d30; (4) training-based VAR acceleration method (CoDe).

\begin{figure*}[t]
\centering
\includegraphics[width=0.85\textwidth]{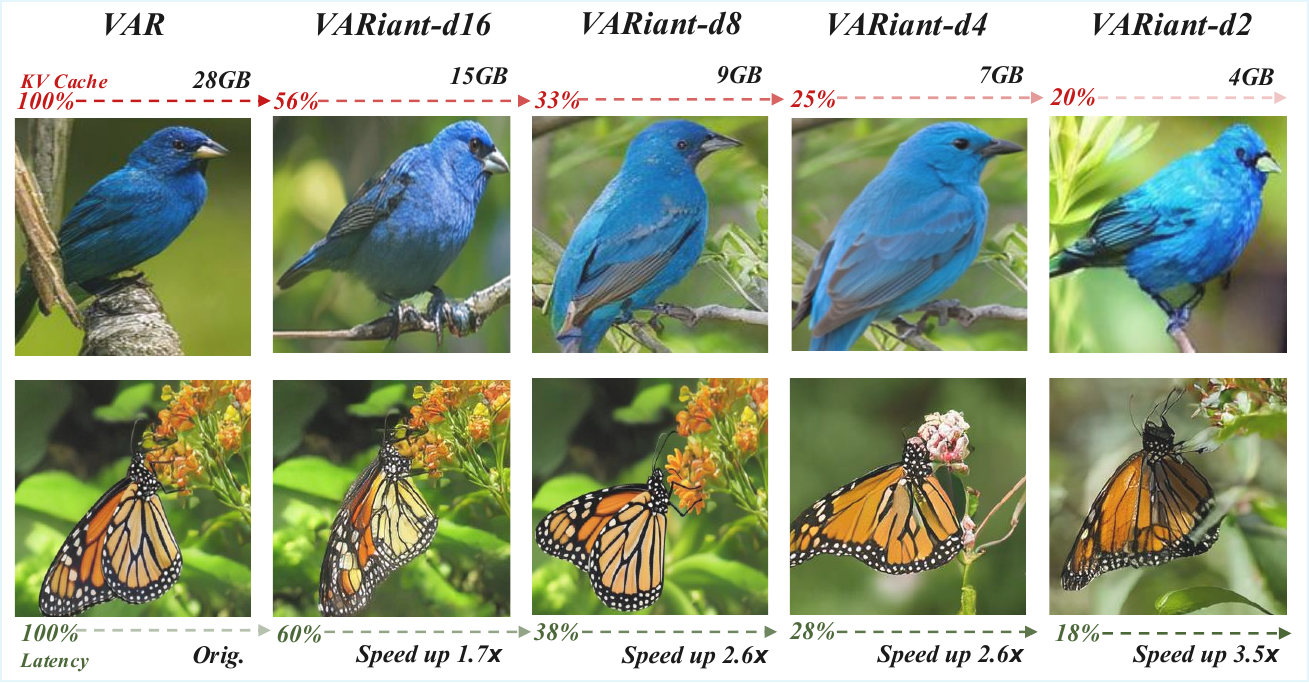}
\caption{Visual quality comparison across different depth configurations. 
All configurations maintain high visual quality with significant memory reduction and inference speedup.}
\label{fig:quality_comparison}
\end{figure*}

\subsection{Main Results}

Table\,\ref{tab:main_results} presents a comprehensive comparison. We analyze the results in the following.

\textbf{Advantages of VAR Paradigm.} VAR reduces generation to 10 steps through next-scale prediction, compared to DiT-XL/2's 50-step diffusion sampling and traditional autoregressive models' 100-384 steps, achieving order-of-magnitude acceleration.

\textbf{Flexible Quality-Efficiency Trade-offs.} Our single 2.0B model provides multiple configurations through dynamic depth switching: d16 and d8 approach VAR-d30's optimal quality (FID 2.05/2.15 \emph{vs.} 1.95) while reducing memory by 40-65\%, and d2 achieves 3.5$\times$ speedup and 80\% memory reduction at moderate quality cost (FID 2.67).

\textbf{Advantages over Training-Based Acceleration.} CoDe employs a dual-model architecture (2.0B+0.3B) achieving FID 2.27 (8.2GB). In comparison, our d4 achieves comparable quality (FID 2.30) with less memory (7.2GB), d8 demonstrates significantly better quality (FID 2.15), and d2 substantially reduces memory (5.5GB). Additionally, the single-model architecture eliminates dual-model deployment complexity and supports zero-cost depth switching.

Figure\,\ref{fig:quality_comparison} shows a visual analysis on models of different depths. While greatly reducing KV sizes and increasing inference speeds, VARiant also maintains high visual quality.

\subsection{Efficiency Analysis}

Table\,\ref{tab:memory} presents memory usage across different batch sizes and depth configurations. KV cache is the dominant memory overhead during inference, accounting for 73.6\% of VAR-d30's total memory. By reducing model depth, our method effectively lowers KV cache consumption: at batch size 64, VARiant-d16/d8/d2 reduce KV cache by 40.4\%, 63.5\%, and 80.8\% respectively.

\begin{table}[h]
\centering
\caption{Memory consumption breakdown at different batch sizes and depth configurations. All measurements are conducted on NVIDIA L20 GPU with batch sizes ranging from 64 to 256. All values in MB. OOM indicates out-of-memory errors.}
\label{tab:memory}
\small
\begin{tabular}{l|ccc}
\toprule
\multirow{2}{*}{\textbf{Method}} & \multicolumn{3}{c}{\textbf{Memory Consumption (MB)↓}} \\
\cmidrule(lr){2-4}
                                  & \textbf{KV Cache} & \textbf{Params} & \textbf{Total} \\
\midrule
\multicolumn{4}{l}{\textit{Batch Size = 64}} \\
\midrule
VAR-d30      & 28687 & 8085 & 38977 \\
VARiant-d16 & 16092 & 8085 & 27015 \\
VARiant-d8  &  9465 & 8085 & 20759 \\
VARiant-d2  &  5495 & 8085 & 18870 \\
\midrule
\multicolumn{4}{l}{\textit{Batch Size = 128}} \\
\midrule
VAR-d30      & \multicolumn{3}{c}{OOM} \\
VARiant-d16 & \multicolumn{3}{c}{OOM} \\
VARiant-d8  & 20931 & 8085 & 33397 \\
VARiant-d2  & 10991 & 8085 & 29635 \\
\midrule
\multicolumn{4}{l}{\textit{Batch Size = 256}} \\
\midrule
VAR-d30      & \multicolumn{3}{c}{OOM} \\
VARiant-d16 & \multicolumn{3}{c}{OOM} \\
VARiant-d8  & \multicolumn{3}{c}{OOM} \\
VARiant-d2  & 21982 & 8085 & 40927 \\
\bottomrule
\end{tabular}
\end{table}

This memory compression well improves batch size scalability: VAR-d30 encounters OOM at batch size 128, while VARiant-d8 can stably run batch size 128, and VARiant-d2 even supports batch size 256. This scalability is valuable in practice, allowing users to flexibly select depth configurations based on their requirements, enabling a single model to adapt to diverse deployment scenarios.

\subsection{Ablation Studies}

To validate our core design choices, we conduct ablation studies on: (1) the impact of joint training on subnet performance, and (2) the effectiveness of our bridge zone design.
%

\subsubsection{Impact of Joint Training on Subnet Performance}

We compare the performance of different subnet depths with and without training adaptation. Training-free baselines directly use pretrained VAR-d30 weights. Table\,\ref{tab:ablation_subnet} manifests the results.

\begin{table}[!h]
\centering
\caption{FID comparison (↓) of different subnet depths before and after training. Training-free baselines use pretrained VAR-d30 weights. Each row represents one training configuration.}
\label{tab:ablation_subnet}
\small
\resizebox{\columnwidth}{!}{%
\begin{tabular}{l|cc|cc}
\toprule
\textbf{Subnet Depth} & \multicolumn{2}{c|}{\textbf{Training-free FID↓}} & \multicolumn{2}{c}{\textbf{Joint Training FID↓}} \\
\cmidrule(lr){2-3} \cmidrule(lr){4-5}
& \textbf{Subnet} & \textbf{Full} & \textbf{Subnet} & \textbf{Full} \\
\midrule
$d=2$ (5\%) & 132 & 1.95 & 2.97 & 2.14 \\
$d=4$ (13\%) & 130 & 1.95 & 2.28 & 2.13 \\
$d=8$ (25\%) & 22.7 & 1.95 & 2.12 & 2.02 \\
$d=16$ (50\%) & 5.42 & 1.95 & 2.05 & 1.96 \\
\bottomrule
\end{tabular}%
}
\end{table}

Training-free baselines cause severe degradation: extremely shallow subnets ($d$=2, 4) almost completely collapse (FID$>$130), while deeper subnets also degrade significantly (\emph{e.g.}, $d$=16's FID drops from 1.95 to 5.42). After joint training, all subnets achieve substantial recovery. Extremely shallow subnets recover from failure (FID$>$130) to usable levels (FID 2.28-2.97), achieving over 50$\times$ improvement; deeper subnets ($d$=8, 16) reach FID 2.05-2.12, approaching full network performance.

Joint training results in slight variations in full network FID (1.96-2.14): deeper subnets have higher overlap with the full network and maintain better full network performance, while shallower subnets require larger weight adjustments and lead to slightly lower performance. Despite these variations, all configurations remain within practical ranges, validating the effectiveness of our training strategy.

\subsubsection{Effectiveness of Bridge Zone Design}

\begin{table}[!t]
\centering
\caption{Ablation study on bridge zone design with flexible zone at r7--r10 ($d=16$, $D=30$).}
\label{tab:ablation_bridge}
\small
\begin{tabular}{l|cc|cc}
\toprule
\multirow{2}{*}{\textbf{Design}} & \multicolumn{2}{c|}{\textbf{Subnet (d16)}} & \multicolumn{2}{c}{\textbf{Full Network}} \\
\cmidrule(lr){2-3} \cmidrule(lr){4-5}
& \textbf{FID↓} & \textbf{IS↑} & \textbf{FID↓} & \textbf{IS↑} \\
\midrule
skip $r_1$-$r_{10}$ & 9.44 & 108 & 3.12 & 187 \\
\textbf{skip $r_7$-$r_{10}$} & \textbf{2.05} & \textbf{314} & \textbf{1.96} & \textbf{301} \\
\bottomrule
\end{tabular}
\end{table}

Table\,\ref{tab:ablation_bridge} validates the effectiveness of our bridge zone design. When all scales ($r_1$-$r_{10}$) skip layers, the skipped layers receive no gradients during training, causing gradient starvation and training conflicts that severely degrade performance (subnet FID 9.44, full network FID 3.12). In contrast, our design uses full depth for early scales ($r_1$-$r_6$) and subnet depth for later scales ($r_7$-$r_{10}$), establishing a ``gradient bridge'' mechanism that ensures all layers continuously receive gradient updates, achieving substantial performance improvements (subnet FID 2.05, full network FID 1.96).

\begin{figure*}[t]
\centering
\includegraphics[width=\textwidth]{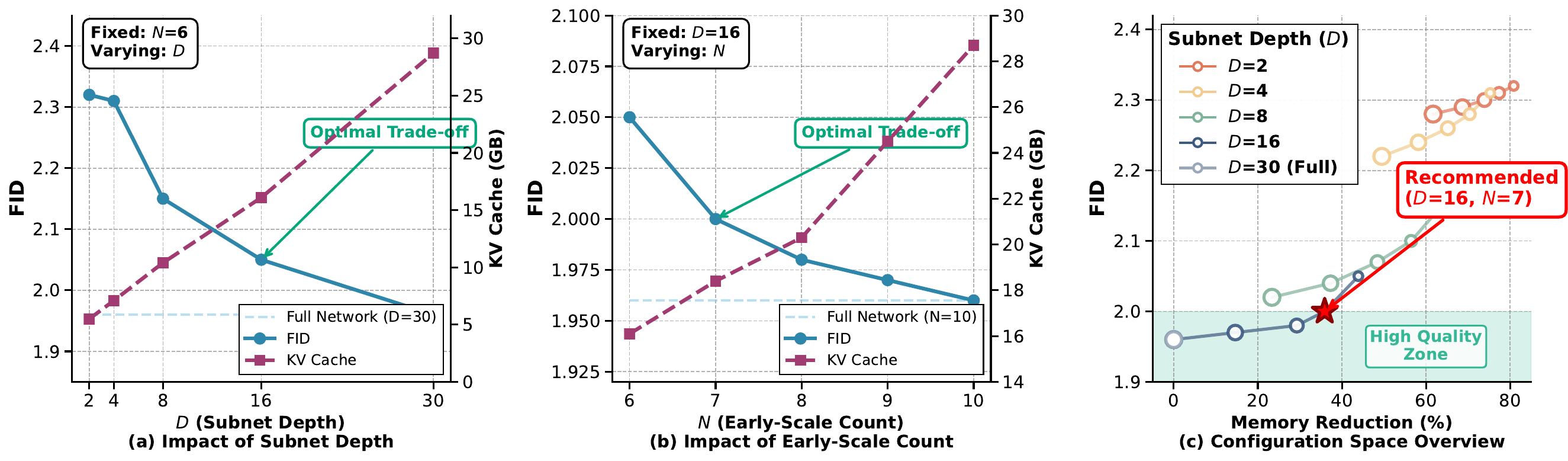}
\caption{Configuration parameter analysis. (a) Impact of subnet depth $D$ (fixed $N$=6): quality \emph{vs.} memory trade-off. (b) Impact of early-scale count $N$ (fixed $D$=16): diminishing returns with increasing $N$. (c) Configuration space: colored trajectories for different $D$ values, marker size indicates $N$. Red star: recommended configuration ($D$=16, $N$=7) with FID 2.00 and 36\% memory reduction.}
\label{fig:ablation_D_N}
\end{figure*}

\subsection{Configuration Analysis}

Our method can achieve flexible quality-efficiency trade-offs through two hyperparameters: subnet depth $D$ controls the number of subnet layers, thereby determining the lower bound of computational capacity and memory footprint; early-scale count $N$ controls at which scale to start using subnet depth, further determining the generation refinement level.

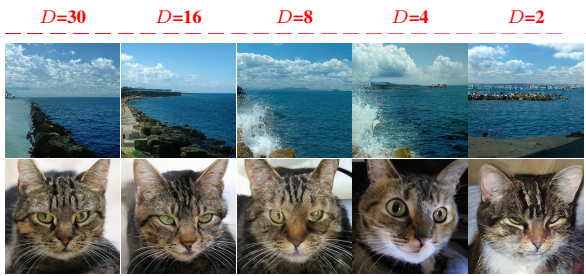
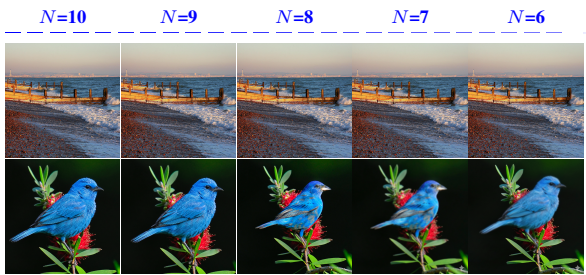
\begin{figure}[h]
\centering

\begin{subfigure}[t]{\columnwidth}
\centering
\begin{tikzpicture}
    \def\imgwidth{1.52cm}
    \def\imgheight{1.52cm}
    
    \node[inner sep=0pt] at (0, 0) 
        {\includegraphics[width=\imgwidth, height=\imgheight]{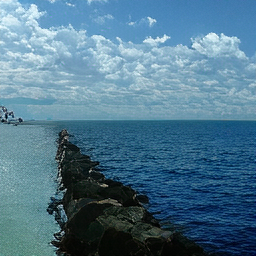}};
    \node[inner sep=0pt] at (1.54cm, 0) 
        {\includegraphics[width=\imgwidth, height=\imgheight]{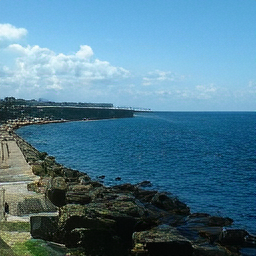}};
    \node[inner sep=0pt] at (3.08cm, 0) 
        {\includegraphics[width=\imgwidth, height=\imgheight]{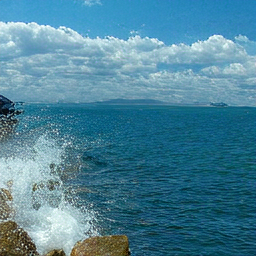}};
    \node[inner sep=0pt] at (4.62cm, 0) 
        {\includegraphics[width=\imgwidth, height=\imgheight]{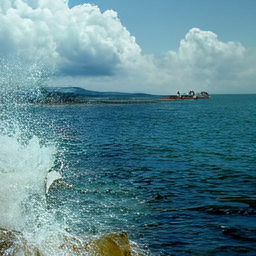}};
    \node[inner sep=0pt] at (6.16cm, 0) 
        {\includegraphics[width=\imgwidth, height=\imgheight]{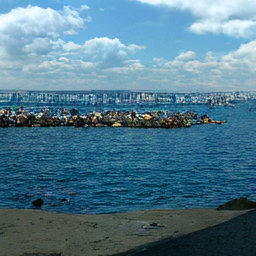}};
    
    \node[inner sep=0pt] at (0, -1.54cm) 
        {\includegraphics[width=\imgwidth, height=\imgheight]{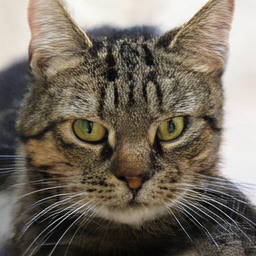}};
    \node[inner sep=0pt] at (1.54cm, -1.54cm) 
        {\includegraphics[width=\imgwidth, height=\imgheight]{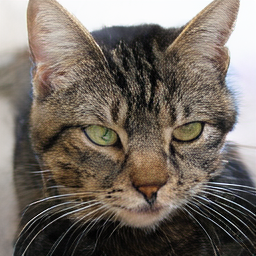}};
    \node[inner sep=0pt] at (3.08cm, -1.54cm) 
        {\includegraphics[width=\imgwidth, height=\imgheight]{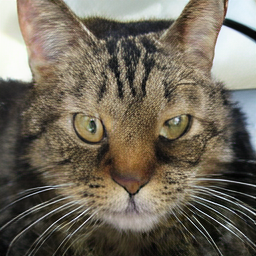}};
    \node[inner sep=0pt] at (4.62cm, -1.54cm) 
        {\includegraphics[width=\imgwidth, height=\imgheight]{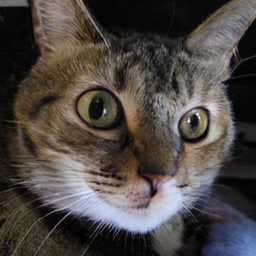}};
    \node[inner sep=0pt] at (6.16cm, -1.54cm) 
        {\includegraphics[width=\imgwidth, height=\imgheight]{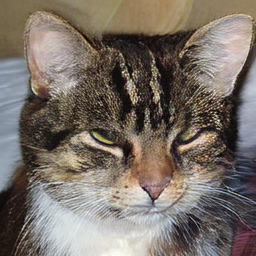}};
    
    \def\arrowY{0.9cm}
    \pgfmathsetmacro{\arrowStart}{-0.76}
    \pgfmathsetmacro{\arrowEnd}{6.92}
    
    \foreach \i in {0,1,...,24} {
        \pgfmathsetmacro{\xstart}{\arrowStart+0.3*\i}
        \pgfmathsetmacro{\xend}{\arrowStart+0.3*\i+0.2}
        \pgfmathsetmacro{\intensity}{100-2*\i}
        \fill[red!\intensity] (\xstart, \arrowY-0.06) rectangle (\xend, \arrowY+0.06);
    }
    
    \fill[red!50] (\arrowEnd, \arrowY-0.06) -- (\arrowEnd+0.2, \arrowY) -- (\arrowEnd, \arrowY+0.06) -- cycle;
    
    \node[above, font=\scriptsize\bfseries, red] at (0, \arrowY+0.2) {$D$=30};
    \node[above, font=\scriptsize\bfseries, red] at (1.54, \arrowY+0.2) {$D$=16};
    \node[above, font=\scriptsize\bfseries, red] at (3.08, \arrowY+0.2) {$D$=8};
    \node[above, font=\scriptsize\bfseries, red] at (4.62, \arrowY+0.2) {$D$=4};
    \node[above, font=\scriptsize\bfseries, red] at (6.16, \arrowY+0.2) {$D$=2};
\end{tikzpicture}
\subcaption{Subnet depth ($D$) reduction from 30 to 2 layers}
\label{fig:depth_reduction}
\end{subfigure}

\vspace{0.4cm}

\begin{subfigure}[t]{\columnwidth}
\centering
\begin{tikzpicture}
    \def\imgwidth{1.52cm}
    \def\imgheight{1.52cm}
    
    \node[inner sep=0pt] at (0, 0) 
        {\includegraphics[width=\imgwidth, height=\imgheight]{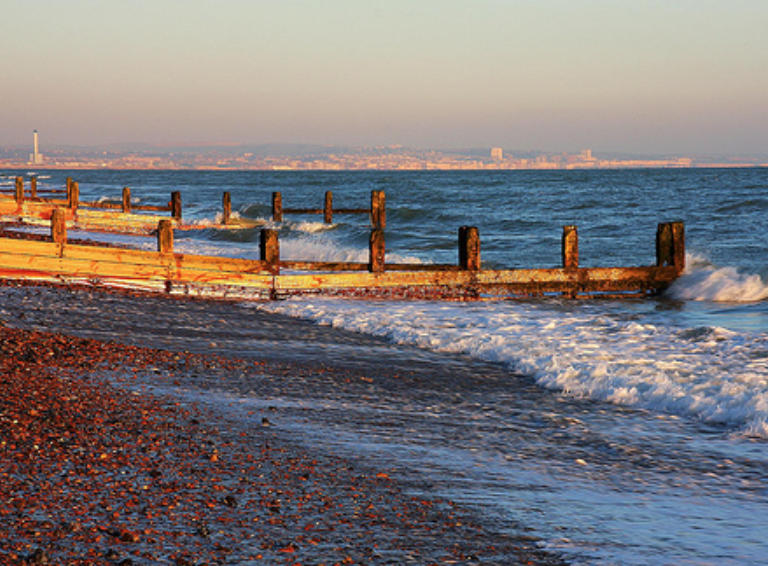}};
    \node[inner sep=0pt] at (1.54cm, 0) 
        {\includegraphics[width=\imgwidth, height=\imgheight]{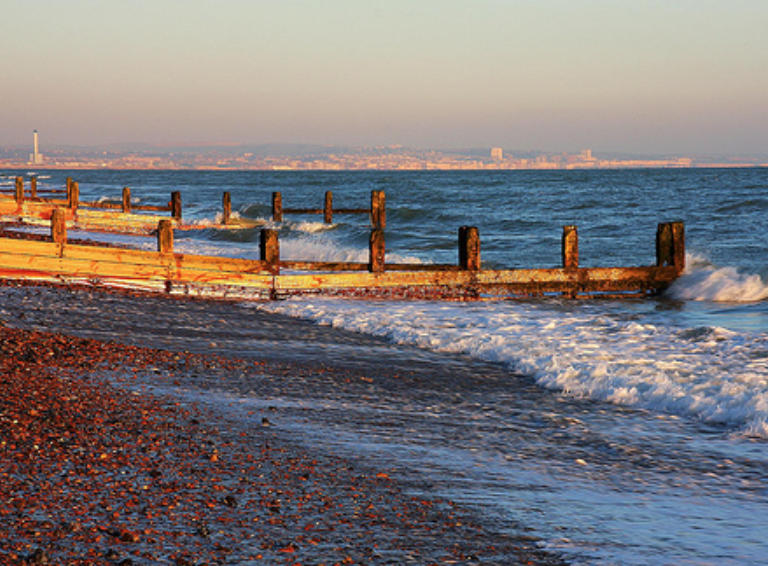}};
    \node[inner sep=0pt] at (3.08cm, 0) 
        {\includegraphics[width=\imgwidth, height=\imgheight]{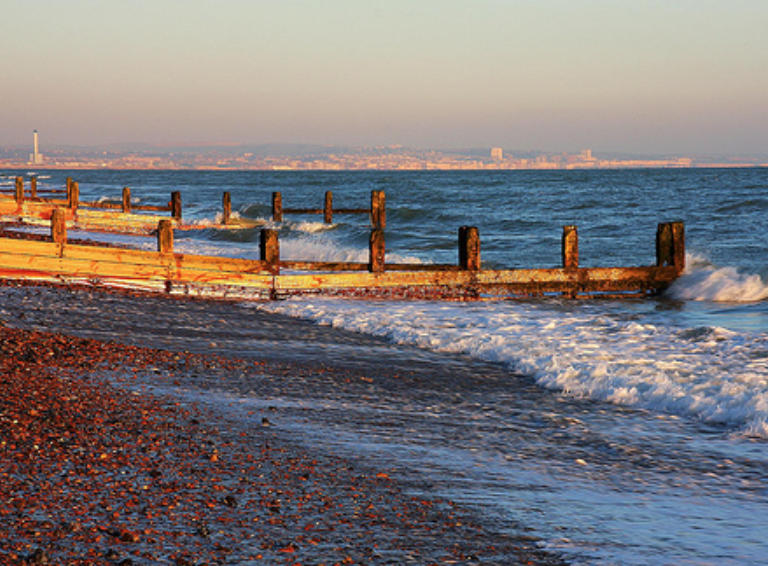}};
    \node[inner sep=0pt] at (4.62cm, 0) 
        {\includegraphics[width=\imgwidth, height=\imgheight]{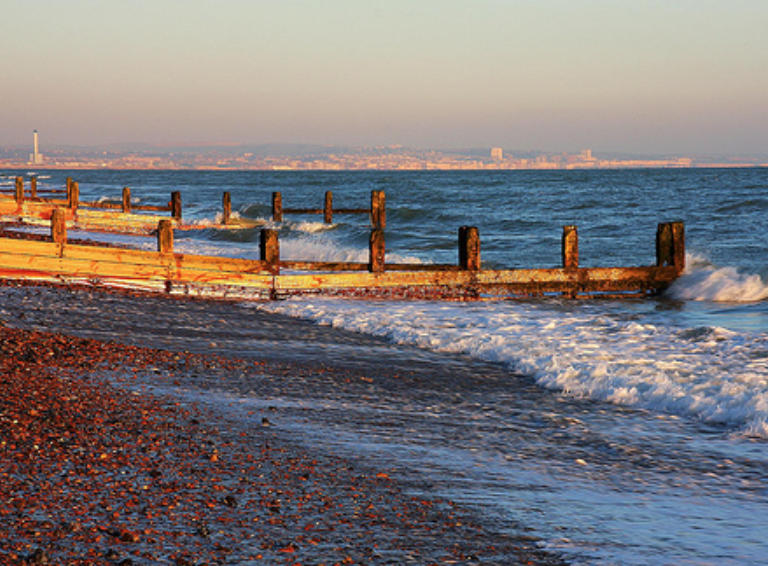}};
    \node[inner sep=0pt] at (6.16cm, 0) 
        {\includegraphics[width=\imgwidth, height=\imgheight]{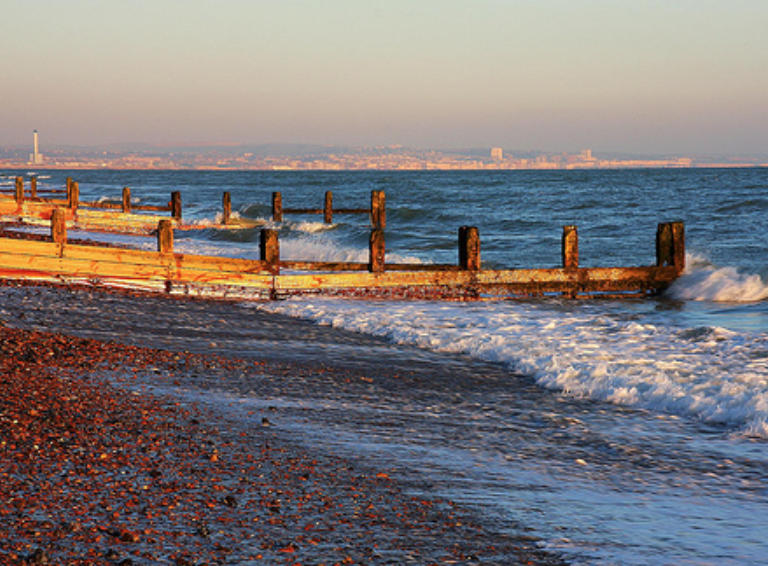}};
    
    \node[inner sep=0pt] at (0, -1.54cm) 
        {\includegraphics[width=\imgwidth, height=\imgheight]{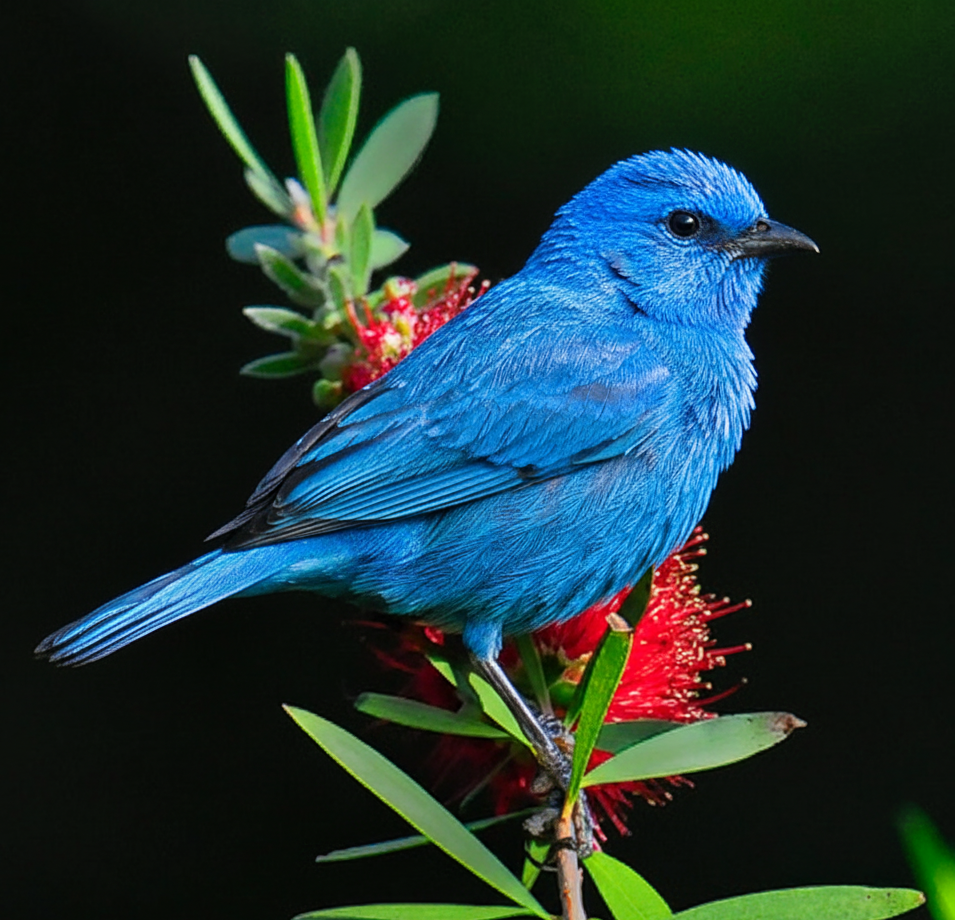}};
    \node[inner sep=0pt] at (1.54cm, -1.54cm) 
        {\includegraphics[width=\imgwidth, height=\imgheight]{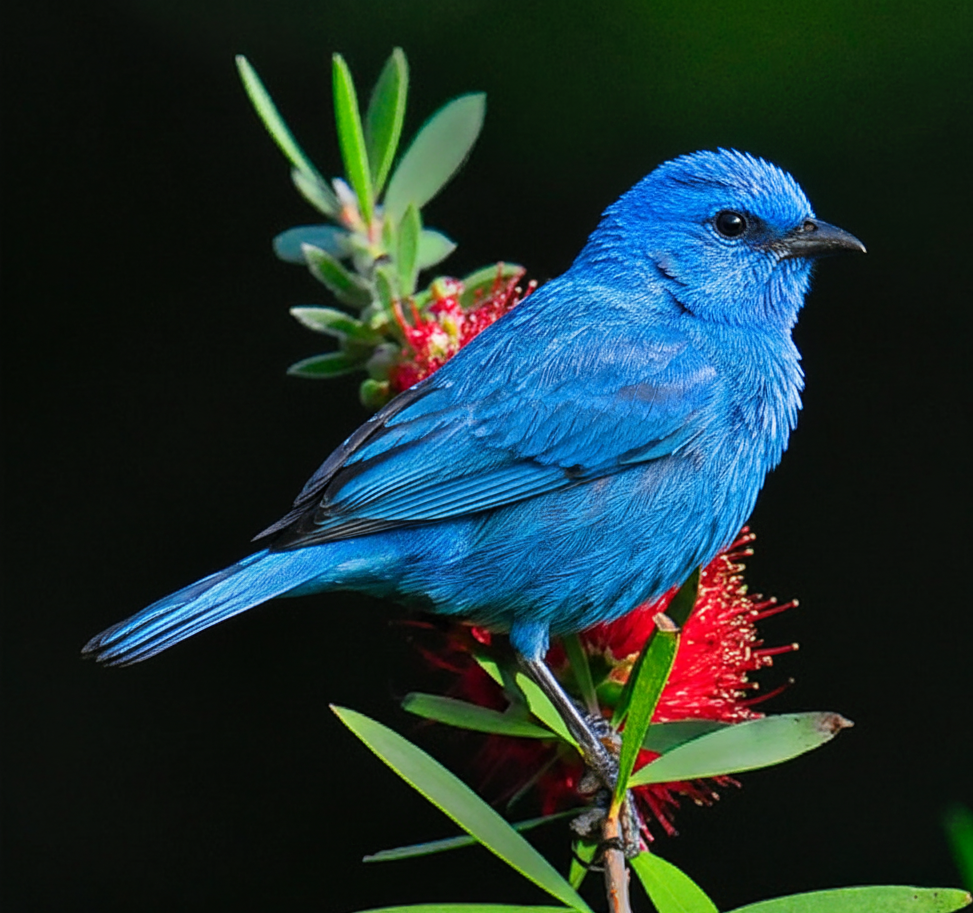}};
    \node[inner sep=0pt] at (3.08cm, -1.54cm) 
        {\includegraphics[width=\imgwidth, height=\imgheight]{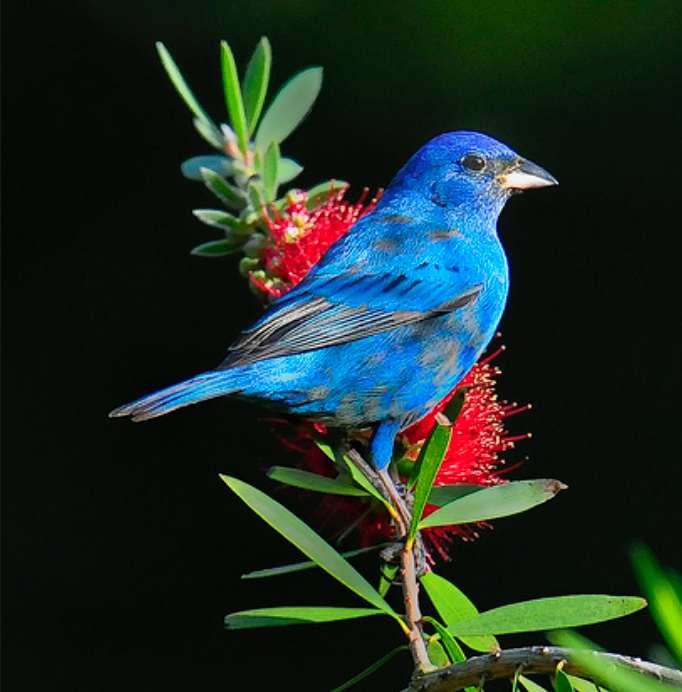}};
    \node[inner sep=0pt] at (4.62cm, -1.54cm) 
        {\includegraphics[width=\imgwidth, height=\imgheight]{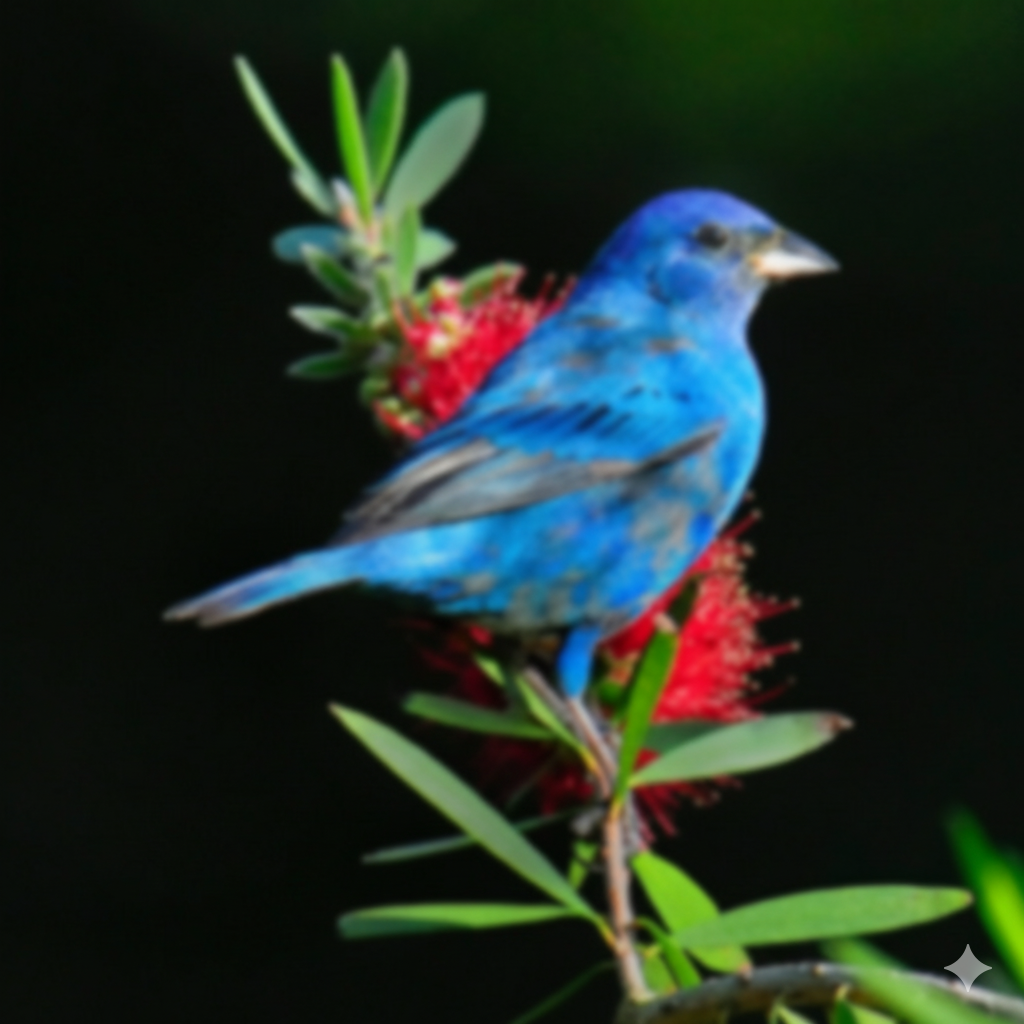}};
    \node[inner sep=0pt] at (6.16cm, -1.54cm) 
        {\includegraphics[width=\imgwidth, height=\imgheight]{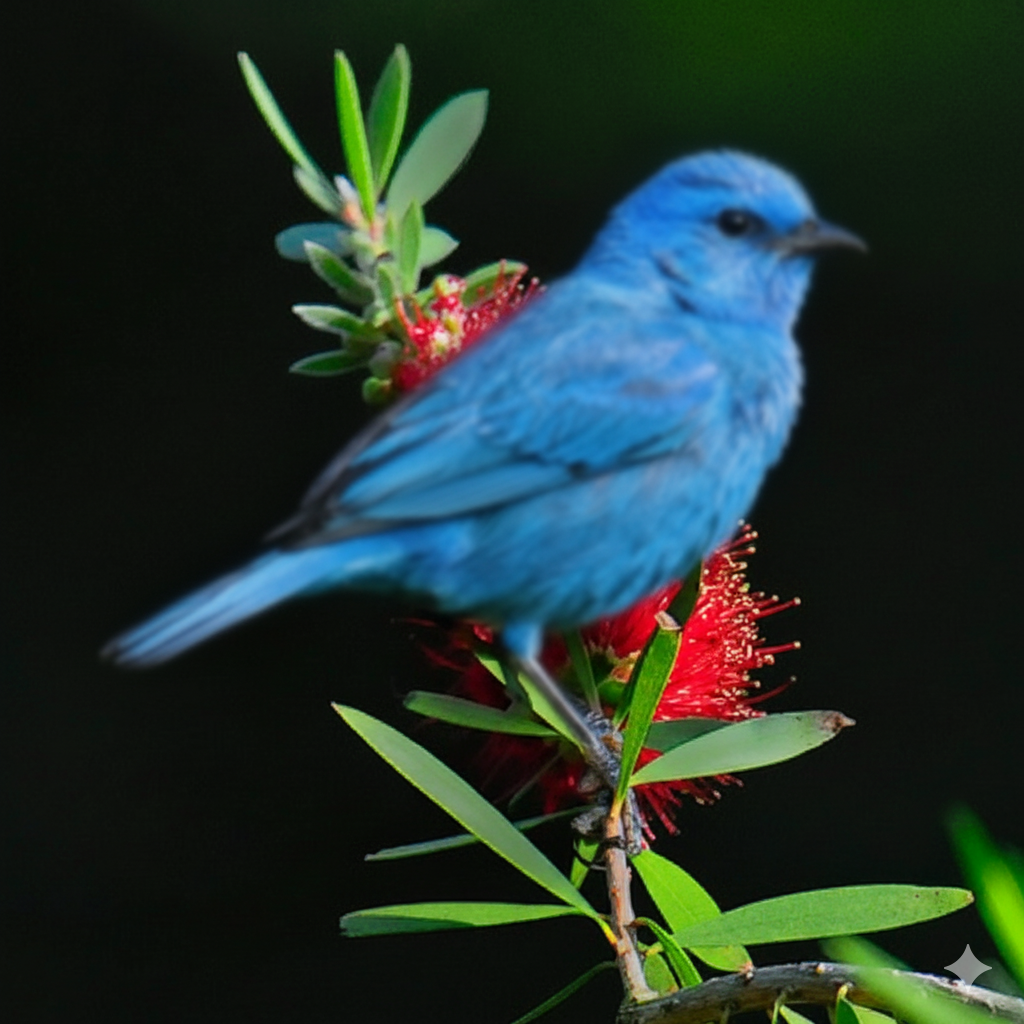}};
    
    \def\arrowY{0.9cm}
    \pgfmathsetmacro{\arrowStart}{-0.76}
    \pgfmathsetmacro{\arrowEnd}{6.92}
    
    \foreach \i in {0,1,...,24} {
        \pgfmathsetmacro{\xstart}{\arrowStart+0.3*\i}
        \pgfmathsetmacro{\xend}{\arrowStart+0.3*\i+0.2}
        \pgfmathsetmacro{\intensity}{100-2*\i}
        \fill[blue!\intensity] (\xstart, \arrowY-0.06) rectangle (\xend, \arrowY+0.06);
    }
    
    \fill[blue!50] (\arrowEnd, \arrowY-0.06) -- (\arrowEnd+0.2, \arrowY) -- (\arrowEnd, \arrowY+0.06) -- cycle;
    
    \node[above, font=\scriptsize\bfseries, blue] at (0, \arrowY+0.2) {$N$=10};
    \node[above, font=\scriptsize\bfseries, blue] at (1.54, \arrowY+0.2) {$N$=9};
    \node[above, font=\scriptsize\bfseries, blue] at (3.08, \arrowY+0.2) {$N$=8};
    \node[above, font=\scriptsize\bfseries, blue] at (4.62, \arrowY+0.2) {$N$=7};
    \node[above, font=\scriptsize\bfseries, blue] at (6.16, \arrowY+0.2) {$N$=6};
\end{tikzpicture}
\subcaption{Early-scale count ($N$) reduction from 10 to 6}
\label{fig:scales_reduction}
\end{subfigure}
\caption{Visual results of progressive configuration adjustment. (a) Subnet depth $D$ from 30 to 2 layers. (b) Early-scale count $N$ from 10 to 6, showing quality-efficiency trade-offs.}
\label{fig:visual_comparison}
\end{figure}

Figure\,\ref{fig:visual_comparison} presents qualitative visual comparison across configurations. We now systematically analyze configuration selection strategies through Figure\,\ref{fig:ablation_D_N}.

\subsubsection{Impact of Subnet Depth ($D$)}

While previous sections demonstrated absolute performance of different depth configurations, here we analyze the trade-off process for configuration selection. Figure\,\ref{fig:ablation_D_N}(a)'s dual-axis plot reveals the quality-memory trade-off: quality improves continuously with depth but with diminishing returns, while memory exhibits approximately linear growth. $D$=16 reaches the optimal balance point (green annotation), maintaining near-optimal quality while achieving significant memory reduction, suitable for most application scenarios.

\subsubsection{Impact of Early-Scale Count ($N$)}

With $D$=16 fixed, Figure\,\ref{fig:ablation_D_N}(b) shows a diminishing returns curve as $N$ increases from 6 to 10. Increasing $N$ from 6 to 7 yields the optimal quality-memory ratio (FID improves 0.05 with only 14\% memory increase); further increases from 7 to 10 show diminishing returns, with progressively smaller FID improvements (0.02, 0.01, 0.01) while memory continues growing linearly.

\subsubsection{Coordinated Control of $D$ and $N$}

Figure\,\ref{fig:ablation_D_N}(c) provides a comprehensive configuration space view, with color indicating $D$ and marker size indicating $N$. Analysis reveals a hierarchical configuration strategy: $D$ serves as the primary control dimension defining performance-efficiency tiers, while $N$ enables fine-grained tuning within each $D$ tier. At $D$=16 and $N$=7 (red star), the configuration achieves near-optimal quality (FID 2.00) with 36\% memory reduction, which is our recommended optimal trade-off point.

In practical deployment, first select $D$ tier based on scenario: $D$=2 or 4 for extreme efficiency, $D$=16 for higher quality requirements; then dynamically adjust $N$ based on actual memory budget. Both parameters support zero-cost runtime switching.

\section{Conclusion}

We introduced VARiant, a unified supernet framework enabling flexible depth adjustment for Visual Autoregressive models through parameter sharing. By exploiting the scale-depth asymmetric dependency, our VARiant allocates full depth to early scales and adaptive shallow subnets to later scales, achieving significant memory reduction and inference acceleration. The dynamic-ratio progressive training strategy effectively resolves optimization conflicts, enabling both subnet and full network to achieve near-optimal performance within a single model. Extensive experiments on ImageNet show that our VARiant breaks through the Pareto frontier of fixed-ratio training, providing flexible quality-efficiency trade-offs for diverse deployments.

\textbf{Limitations and Future Work.} 
We currently train one subnet with the full network. Future work could extend to simultaneously training multiple subnets (\emph{e.g.}, d4/d8/d16/d30). Also, the transition epochs in our three-phase training strategy are currently determined empirically, and developing principled methods to automatically determine optimal phase boundaries could improve training efficiency. Lastly, exploring our scale-aware depth adaptation strategy in other multi-scale generation models represents a promising research direction.

\section{Supplementary Material}
\label{sec:supplementary}

This section provides additional technical details, complete experimental configurations, and extended ablation studies to support the main paper.

\subsection{Algorithm Details}
\label{sec:supp_algorithm}

\subsubsection{Progressive Supernet Training Algorithm}

Algorithm\,\ref{alg:training} presents the complete pseudocode for our three-phase progressive supernet training strategy.

\begin{algorithm}[h]
\caption{Progressive Supernet Training}
\label{alg:training}
\begin{algorithmic}[1]
\REQUIRE Dataset $\mathcal{D}$, Full depth $D=30$, Subnet depth $d$, Phase epochs $(E_1, E_2, E_3)$
\ENSURE Trained supernet $\theta$

\STATE \textbf{// Phase 1: Joint Training (Epochs $0$ to $E_1$)}
\FOR{$\text{epoch} = 0$ \TO $E_1$}
    \FOR{each batch in $\mathcal{D}$}
        \IF{$\text{random}() < 0.2$}
            \STATE Use subnet layers $\mathcal{I}_d$ \big(Eq.\,(\ref{eq:selection}) and Eq.\,(\ref{eq:active}) in main paper\big)
        \ELSE
            \STATE Use full depth $D$
        \ENDIF
        \STATE Compute loss $\mathcal{L}$ (Eq.\,(\ref{eq:phase1_loss}) in main paper) and update $\theta$
    \ENDFOR
\ENDFOR

\STATE \textbf{// Phase 2: Progressive Transition (Epochs $E_1+1$ to $E_2$)}
\FOR{$\text{epoch} = E_1+1$ \TO $E_2$}
    \STATE $p = 0.2 + 0.8 \times \frac{\text{epoch}-E_1}{E_2-E_1}$ \quad // Linear increase
    \FOR{each batch in $\mathcal{D}$}
        \IF{$\text{random}() < p$}
            \STATE Use subnet layers $\mathcal{I}_d$
        \ELSE
            \STATE Use full depth $D$
        \ENDIF
        \STATE Compute loss $\mathcal{L}$ and update $\theta$
    \ENDFOR
\ENDFOR

\STATE \textbf{// Phase 3: Subnet Refinement (Epochs $E_2+1$ to $E_3$)}
\FOR{$\text{epoch} = E_2+1$ \TO $E_3$}
    \FOR{each batch in $\mathcal{D}$}
        \STATE Use subnet layers $\mathcal{I}_d$ only
        \STATE Compute loss $\mathcal{L}$ and update $\theta$
    \ENDFOR
\ENDFOR
\end{algorithmic}
\end{algorithm}

\subsubsection{Inference Algorithm}

Algorithm\,\ref{alg:inference} details our inference procedure with dynamic depth switching.

\begin{algorithm}[h]
\caption{Inference with Dynamic Depth Switching}
\label{alg:inference}
\begin{algorithmic}[1]
\REQUIRE Class label $y$, Subnet depth $d$, Bridge zone size $N$
\ENSURE Generated image

\STATE // Compute active layer indices via equidistant sampling
\STATE $\mathcal{I}_d \leftarrow \text{Equidistant\_Sample}(d, D=30)$ \quad // Eq.\,(\ref{eq:selection}) in main paper

\FOR{scale $k = 1$ \TO $K$}
    \IF{$k \leq N$}
        \STATE $\text{layers} \leftarrow \{0, 1, \ldots, D-1\}$ \quad // Full depth (Bridge Zone)
    \ELSE
        \STATE $\text{layers} \leftarrow \mathcal{I}_d$ \quad // Subnet depth (Flexible Zone)
    \ENDIF
    \STATE $r_k \leftarrow \text{Transformer}(r_{<k}, \text{layers})$
\ENDFOR

\RETURN $\text{VQVAE\_Decode}(r_{1:K})$
\end{algorithmic}
\end{algorithm}

\subsubsection{Complete Training Configuration}

Table\,\ref{tab:hyperparameters} and Table\,\ref{tab:phase_duration} provide comprehensive training configurations used in our experiments. Table\,\ref{tab:hyperparameters} lists all hyperparameters including optimizer settings, data augmentation strategies, and hardware specifications. Table\,\ref{tab:phase_duration} details the phase duration configuration for different subnet depths, showing that shallower subnets require longer refinement phases to achieve convergence.

\begin{table}[h]
\centering
\caption{Complete training hyperparameters.}
\label{tab:hyperparameters}
\begin{tabular}{lp{4.5cm}}  
\toprule
\textbf{Configuration} & \textbf{Value} \\
\midrule
Optimizer & AdamW \\
Learning Rate & $1 \times 10^{-6}$ (constant) \\
Weight Decay & 0.05 \\
Batch Size & 1024 (128 per GPU $\times$ 8 GPUs) \\
Warmup & None (finetune from pretrained VAR-d30) \\
Gradient Clipping & 1.0 \\
Mixed Precision & FP16 \\
Hardware & 8$\times$ NVIDIA H100 (80GB) \\
\midrule
\multirow{3}{*}{Data Augmentation} & Random horizontal flip ($p=0.5$) \\
 & RandAugment (2 ops., mag. 9) \\  
 & Mixup ($\alpha=0.2$) \\
\midrule
Bridge Zone Config. & $N=6$ ($r_1$--$r_6$ full depth, \newline $r_7$--$r_{10}$ subnet depth) \\  
\bottomrule
\end{tabular}
\end{table}

\begin{table}[h]
\centering
\caption{Training phase configuration for different subnet depths.}
\label{tab:phase_duration}
\begin{tabular}{lcccc}
\toprule
\textbf{Depth} & \textbf{Phase 1} & \textbf{Phase 2} & \textbf{Phase 3} & \textbf{Total} \\
\midrule
$d=16$ & 5 epochs & 15 epochs & 5 epochs & 25 epochs \\
$d=8$ & 5 epochs & 15 epochs & 8 epochs & 28 epochs \\
$d=4$ & 5 epochs & 15 epochs & 12 epochs & 32 epochs \\
$d=2$ & 5 epochs & 15 epochs & 15 epochs & 35 epochs \\
\bottomrule
\end{tabular}
\vspace{0.3cm}

\end{table}

\subsection{Training Process Visualization}
\label{sec:supp_training_vis}

\begin{figure*}[h]
\centering
\begin{tabular}{@{}ccc@{}}
\includegraphics[width=0.30\textwidth]{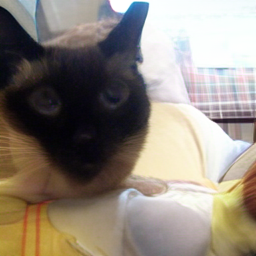} &
\includegraphics[width=0.30\textwidth]{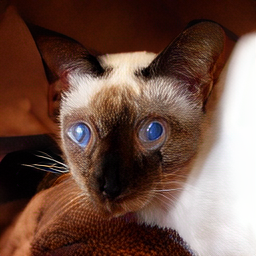} &
\includegraphics[width=0.30\textwidth]{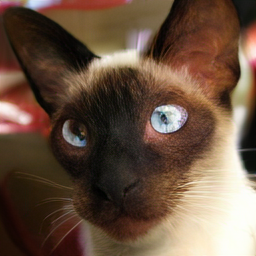} \\
\includegraphics[width=0.30\textwidth]{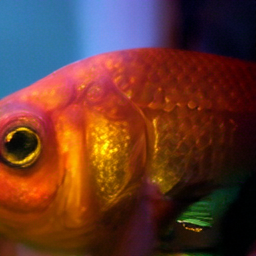} &
\includegraphics[width=0.30\textwidth]{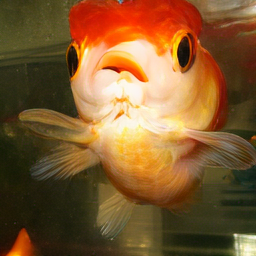} &
\includegraphics[width=0.30\textwidth]{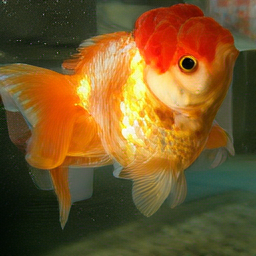} \\
\small{(a) Phase 1 End} & \small{(b) Phase 2 End} & \small{(c) Phase 3 End} \\
\small{Epoch 5} & \small{Epoch 20} & \small{Epoch 25} \\
\end{tabular}
\caption{Subnet generation quality evolution during progressive training ($d=16$). Top row: Cat; Bottom row: Fish. Results demonstrate subnet-only inference across three training phase endpoints.}
\label{fig:training_evolution}
\end{figure*}

Figure\,\ref{fig:training_evolution} visualizes the subnet generation quality evolution ($d=16$) for two representative ImageNet classes throughout our progressive training process. All images are generated using the 16-layer subnet configuration to demonstrate the effectiveness of our training strategy.

\textbf{(a) Phase 1 End (Epoch 5).} After joint training with 20\% subnet sampling probability, the subnet generates recognizable but somewhat blurry images for both cat and fish classes. At this early stage, the subnet has learned basic visual patterns from the full network through knowledge distillation, but fine details remain underdeveloped. The cat's facial features and the fish's body structure are visible but lack sharpness.

\textbf{(b) Phase 2 End (Epoch 20).} Following progressive transition where subnet sampling probability linearly increases to 100\%, generation quality significantly improves. The cat image now exhibits clearer fur texture and more defined facial features, while the fish displays better color saturation and fin details. This demonstrates that the gradual increase in subnet sampling enables smooth adaptation without catastrophic forgetting of learned representations.

\textbf{(c) Phase 3 End (Epoch 25).} After subnet-focused refinement, the final model achieves high-quality generation comparable to the full network baseline. The cat image shows rich texture details with natural lighting and realistic fur patterns, while the fish exhibits vibrant colors and well-defined anatomical structures. These results validate that our three-phase progressive training strategy successfully optimizes the lightweight subnet to match full network quality, achieving an optimal quality-efficiency trade-off for practical deployment.

\subsection{More Ablation Studies: Complete Configuration Space}
\label{sec:supp_ablation}

\begin{table}[h]
\centering
\caption{Complete $(d, N)$ configuration space -- FID scores on ImageNet 256$\times$256.}
\label{tab:config_space}
\begin{tabular}{lccccc}
\toprule
\textbf{$d$} & $N=6$ & $N=7$ & $N=8$ & $N=9$ & $N=10$ \\
\midrule
$d=2$ & 2.97 & 2.89 & 2.84 & 2.80 & 2.78 \\
$d=4$ & 2.28 & 2.23 & 2.20 & 2.18 & 2.16 \\
$d=8$ & 2.12 & 2.08 & 2.05 & 2.03 & 2.02 \\
$d=16$ & 2.05 & \textbf{2.00} & 1.98 & 1.97 & 1.96 \\
$d=30$ & 1.96 & 1.96 & 1.96 & 1.96 & 1.96 \\
\bottomrule
\end{tabular}
\end{table}

Table\,\ref{tab:config_space} presents a comprehensive ablation study over the complete configuration space of subnet depth $d$ and bridge zone size $N$. The results demonstrate that generation quality improves with both larger subnet depth and larger bridge zone size, but with diminishing returns. 

\textbf{Effect of Subnet Depth $d$.} As subnet depth increases from $d=2$ to $d=16$, FID scores consistently improve across all bridge zone configurations. The full network ($d=30$) achieves FID 1.96 as the performance upper bound. Notably, $d=16$ with $N=7$ achieves FID 2.00, only $\Delta=+0.04$ worse than the full network.

\textbf{Effect of Bridge Zone Size $N$.} Larger bridge zones provide more full-depth computation for early scales, improving quality across all subnet depths. However, the improvement saturates beyond $N=8$, indicating that excessive bridge zones offer limited benefits while reducing computational savings.

\textbf{Recommended Configuration.} We recommend $d=16, N=7$ (bold) as the optimal quality-efficiency trade-off. This configuration achieves: (1) FID 2.00 with only 2\% quality degradation compared to the full network; (2) Memory reduction from 28.7GB to 18.4GB ($-36\%$); (3) Inference speedup of $1.7\times$. This balance makes it suitable for practical deployment scenarios requiring both high-quality generation and computational efficiency.

\clearpage
{
    \small
    \bibliographystyle{ieeenat_fullname}
    \bibliography{main}
}


\end{document}